\documentclass{article}
\PassOptionsToPackage{switch}{lineno}
\usepackage{ijcai26}

\usepackage{times}
\usepackage{latexsym}
\usepackage{amssymb}
\usepackage{fvextra} 
\usepackage{algorithm}
\usepackage{algorithmic}
\usepackage{multirow} 
\usepackage{graphicx}    
\usepackage{booktabs}    
\usepackage{amsmath,amssymb} 
\usepackage{enumitem}    
\usepackage{float}       
\usepackage{placeins}  
\usepackage{threeparttable}
\usepackage{makecell}%
\usepackage{soul}
\usepackage{url}
\usepackage[hidelinks]{hyperref}
\usepackage[utf8]{inputenc}
\usepackage[table]{xcolor} 
\usepackage{amsthm}

\title{HiP-LoRA: Budgeted Spectral Plasticity for Robust Low-Rank Adaptation}

\author{
Lixian Chen$^1$
\and
JianHong Tan$^1$\\
\affiliations
$^1$Guangdong University of Technology\\
\emails
3123003175@mail2.gdut.edu.cn, 3123006160@mail2.gdut.edu.cn
}
\begin{document}

\maketitle

\begin{abstract}
Adapting foundation models under resource budgets relies heavily on Parameter-Efficient Fine-Tuning (PEFT), with LoRA being a standard modular solution. However, LoRA suffers from spectral interference. Low-rank updates often concentrate energy on the leading singular directions of pretrained weights, perturbing general capabilities and causing catastrophic forgetting and fragile multi-adapter merging. To resolve this, we propose HiP-LoRA, a spectrum-aware adaptation framework. Utilizing the cached singular value decomposition (SVD) of pretrained layers, HiP-LoRA decomposes updates into two channels: a principal channel within the dominant singular subspace, and a residual low-rank channel in the orthogonal complement. A singular-value-weighted stability budget on the principal channel continuously balances pretrained behavior preservation with task-specific plasticity. Experiments on Llama-3.1-8B demonstrate that under matched budgets, HiP-LoRA drastically reduces pretraining degradation and multi-adapter MergeFail, robustly outperforming baselines in interference-sensitive tasks like continual tuning and knowledge editing.
\end{abstract}

\begin{figure}[t]
  \centering
  \includegraphics[width=\linewidth, trim=0cm 4cm 0cm 4cm, clip]{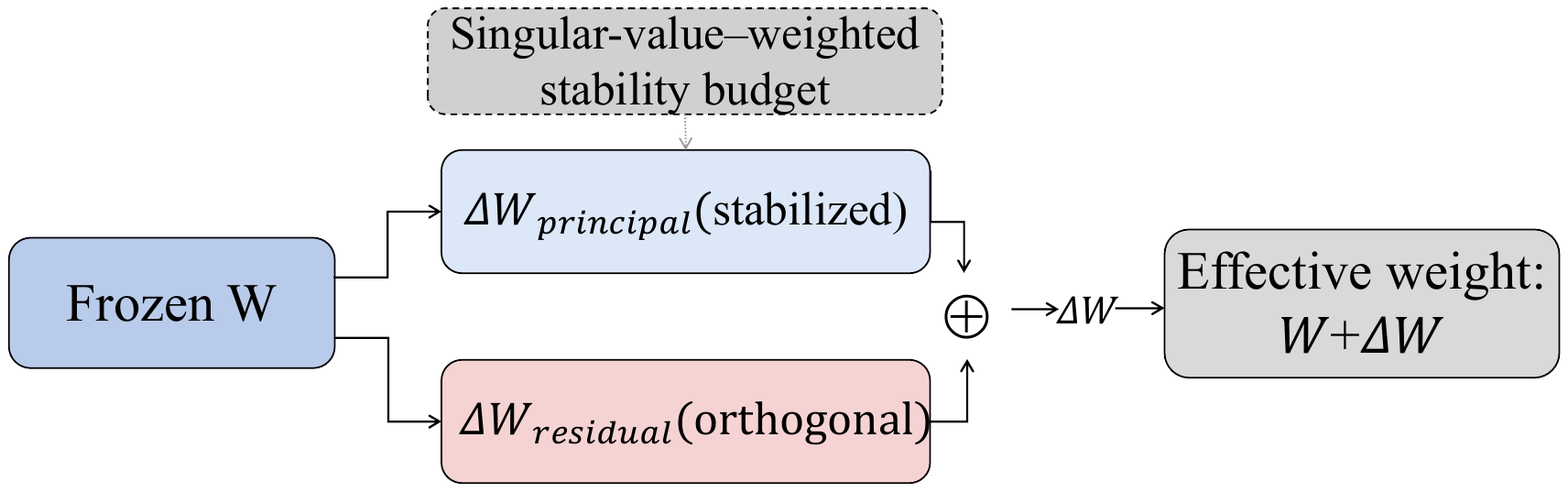}
  \caption{\textbf{Motivation of HiP-LoRA.}
Although LoRA updates are low-rank, their drift can still concentrate on dominant singular directions of the frozen pretrained weight, which may harm pretrained capabilities and reduce robustness under continual adaptation and adapter merging.
HiP-LoRA addresses this by decomposing each update into a principal component regulated by a singular-value--weighted stability budget and a residual component that preserves adaptation capacity.}
  \label{fig:motivation}
\end{figure}

\section{Introduction}
\label{sec:intro}

Foundation models and large pretrained neural backbones are now deployed across a wide range of tasks, including multimodal reasoning, medical imaging, reconstruction, and restoration~\cite{shi2026mmerror,cxh1,cxh2,cxh4,cxh7,cxh9,cxh10}, but practical use often requires repeated adaptation under limited compute, memory, and storage budgets. At the same time, recent evaluation has shown that controlled model behavior and robustness to errors remain important concerns in complex multimodal settings~\cite{shi2026mmerror}. These trends make efficient and reliable adaptation a central problem.

Parameter-efficient finetuning (PEFT)~\cite{houlsby2019parameter} addresses this need by updating a small set of trainable parameters while keeping the pretrained backbone frozen. Representative approaches include prompt tuning~\cite{lester2021power}, bias-only finetuning~\cite{zaken2022bitfit}, and low-rank adaptation methods such as LoRA~\cite{hu2022lora} and QLoRA~\cite{dettmers2023qlora}. In particular, LoRA is attractive because its low-rank updates can be merged into the base model at deployment, which supports modular continual updates and composition of independently trained adapters~\cite{pfeiffer2021adapterfusion,mitchell2021fast,wortsman2022model,ilharco2022editing}. This practical success has made LoRA one of the standard choices for adapting large language models.

However, efficient parameterization alone does not resolve the problem of interference. When a model is adapted repeatedly or when several adapters are combined, the new updates can damage broad pretrained behavior and produce forgetting, negative transfer, or unstable composition~\cite{kirkpatrick2017overcoming,lopez2017gradient,french1999catastrophic}. We study this issue from the spectral structure of pretrained weights. The singular value decomposition of a frozen pretrained matrix orders directions according to their importance in the pretrained parameter space. From this perspective, interference is not determined only by the total size of an update. It also depends on whether the update places substantial energy on dominant singular directions that support broad pretrained behavior. Even a low-rank update can therefore cause large and undesirable functional changes when its drift is concentrated in these dominant directions~\cite{aghajanyan2021intrinsic}.

This observation exposes a gap in existing adaptation methods. Some methods improve optimization or parameterization, while others use dominant subspaces for initialization or enforce orthogonality for preservation. Yet these designs do not directly provide a continuous mechanism that can regulate how much editing is allowed inside dominant pretrained directions while preserving sufficient task plasticity. As a result, low-rank adaptation can still perturb directions that support broad generalization, which leads to degraded retention, brittle adapter composition, and poor edit locality~\cite{yadav2023ties,wang2023orthogonal}.

Motivated by this problem, we propose HiP-LoRA, a spectrum-aware PEFT method inspired by Complementary Learning Systems~\cite{kumaran2016learning,mcclelland1995there}. HiP-LoRA represents each layer update in the cached top singular basis of the pretrained weight matrix and decomposes the update into two coordinated channels. The first channel edits gains inside the dominant singular subspace. The second channel performs residual low-rank adaptation in the two-sided orthogonal complement. A stability budget weighted by singular values then controls the magnitude of edits in dominant directions, which yields a continuous trade-off between preservation of pretrained behavior and task-specific adaptation. Under matched budgets on Llama-3.1-8B, HiP-LoRA reduces Retain from 12.3 to 4.9 percentage points while maintaining or improving downstream task performance, and it substantially improves the robustness of adapter merging without additional training.

The main contributions are listed below:

(1) We study interference in parameter-efficient adaptation from the spectral structure of pretrained weights, and show that low-rank updates can still induce concentrated functional perturbations along leading singular directions. This perspective suggests that retention depends on how update energy is distributed across singular directions, not only on the total update magnitude.

(2) We propose HiP-LoRA, a spectrum-aware low-rank adaptation framework that combines a principal editing channel, a residual complementary channel, and a stability budget weighted by singular values. This design provides continuous control over preservation and adaptation without relying on hard inclusion or exclusion constraints.

(3) We validate HiP-LoRA on single-task adaptation, adapter merging without additional training, continual instruction tuning, and knowledge editing. Across these settings, HiP-LoRA improves retention, merging robustness, and edit locality under matched parameter budgets.

\section{Related Work}

\subsection{Parameter-efficient adaptation}

Parameter-efficient finetuning updates a small subset of parameters while keeping the pretrained backbone fixed, which makes adaptation practical for large models~\cite{houlsby2019parameter}. Representative designs include prompt tuning~\cite{lester2021power}, bias-only finetuning~\cite{zaken2022bitfit}, and low-rank adaptation through LoRA and QLoRA~\cite{hu2022lora,dettmers2023qlora}. Recent surveys summarize a wide range of PEFT strategies, and follow-up work continues to refine the parameterization and optimization of low-rank updates~\cite{han2024parameter,liu2024dora}. These methods improve efficiency and optimization, but they do not explicitly control how update energy is distributed across the spectral coordinates of pretrained weights. Our work focuses on this missing degree of control.

\subsection{Spectral and subspace-aware adaptation}

Several methods use spectral or subspace structure to improve adaptation. Prior work has shown that finetuning often lies in a low-dimensional intrinsic subspace, which motivates structured parameterizations of model updates~\cite{aghajanyan2021intrinsic}. PiSSA uses singular vectors of pretrained weights for initialization and parameterization in order to improve optimization and data efficiency~\cite{meng2024pissa}. Orthogonal or complement-subspace methods instead protect pretrained representations by constraining updates away from important directions~\cite{wang2023orthogonal}. These approaches reveal the value of spectral structure, but they mainly decide where updates can occur. In contrast, HiP-LoRA uses the pretrained singular basis as a fixed coordinate system and continuously regulates how much editing is allowed inside dominant directions while preserving residual adaptation capacity outside them.

\subsection{Adapter composition, merging, and interference}

The modular form of PEFT has encouraged research on combining independently trained adapters and model updates. Adapter fusion and related methods study how multiple task-specific modules can be composed after separate training~\cite{pfeiffer2021adapterfusion}. Other work considers editing, task arithmetic, and direct composition of learned updates in weight space~\cite{mitchell2021fast,wortsman2022model,ilharco2022editing}. Recent studies of model merging further examine alignment, permutation mismatch, and interference across networks or updates~\cite{yadav2023ties,bansal2021revisiting,ainsworth2022git}. These methods address how to combine learned components after training. Our focus is different. We design the adapter update itself so that it is easier to preserve general behavior and more robust to composition from the start.

\subsection{Stability, plasticity, and structured adaptation}

The tension between adaptation and retention is closely related to the long-standing problem of catastrophic forgetting in continual learning~\cite{kirkpatrick2017overcoming,lopez2017gradient,french1999catastrophic}. Beyond language model adaptation, recent work in structured spatio-temporal modeling and domain-specific visual learning has also shown that explicit geometric or architectural constraints can improve stable representation learning and reconstruction quality~\cite{10.1145/3774904.3792090}. At a conceptual level, Complementary Learning Systems provides a useful view of how stable knowledge and fast adaptation can coexist~\cite{kumaran2016learning,mcclelland1995there}. HiP-LoRA draws motivation from this perspective, but turns it into a concrete PEFT mechanism in the spectral basis of pretrained weights, with an explicit budget that controls the balance between retention and adaptation.

\section{Method}
\label{sec:method}

\paragraph{Setup and notation.}
For a frozen pretrained weight matrix $\mathbf{W}\in\mathbb{R}^{m\times n}$, we compute and cache a rank-$k$ truncated SVD
\[
\mathbf{W}\approx \mathbf{U}_k\operatorname{diag}(\boldsymbol{\sigma})\mathbf{V}_k^\top,
\]
where $\mathbf{U}_k\in\mathbb{R}^{m\times k}$, $\mathbf{V}_k\in\mathbb{R}^{n\times k}$, and $\boldsymbol{\sigma}\in\mathbb{R}_+^k$ are fixed throughout adaptation.
We further define the projectors
\[
\mathbf{P}_U=\mathbf{U}_k\mathbf{U}_k^\top,\qquad
\mathbf{P}_V=\mathbf{V}_k\mathbf{V}_k^\top.
\]
Following LoRA, we write a rank-$r$ update as
\[
\Delta\mathbf{W}=s\,\mathbf{B}\mathbf{A},
\]
where $r\ll \min(m,n)$ and $s=\alpha/r$ is the standard LoRA scaling.

\begin{figure*}[!t]
  \centering
  \includegraphics[width=0.95\textwidth, trim=1cm 5.5cm 0cm 4cm, clip]{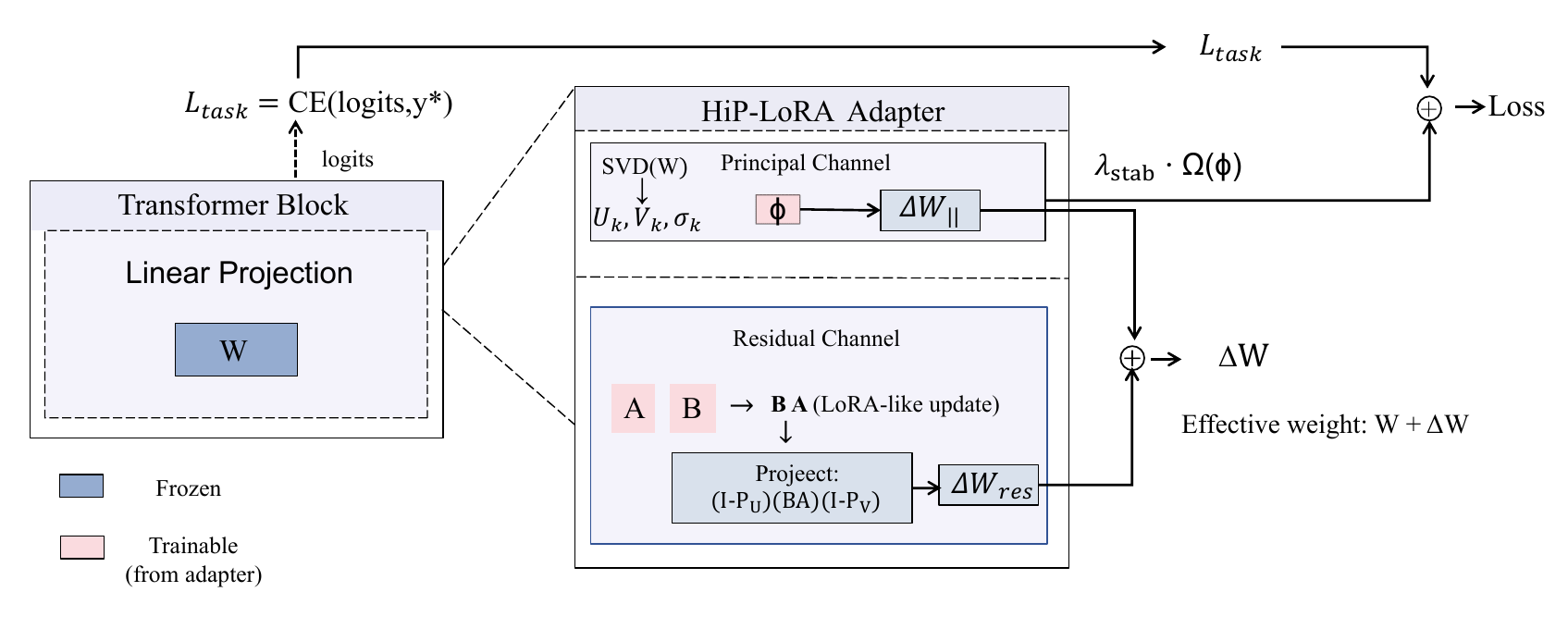}
  \caption{\textbf{HiP-LoRA mechanism.}
Given a frozen transformer projection $\mathbf{W}$, we cache its top-$k$ singular directions
$\mathbf{W}\approx \mathbf{U}_k\operatorname{diag}(\boldsymbol{\sigma})\mathbf{V}_k^\top$ and keep $(\mathbf{U}_k,\mathbf{V}_k,\boldsymbol{\sigma})$ fixed.
HiP-LoRA decomposes the adapter update into two channels:
a principal channel $\mathbf{U}_k\operatorname{diag}(\boldsymbol{\phi})\mathbf{V}_k^\top$ that performs controlled editing along dominant pretrained directions, and
a residual channel $s\,\Delta\mathbf{W}_{\mathrm{res}}$ that applies complementary low-rank adaptation in the two-sided orthogonal complement.
Training minimizes a task loss together with a singular-value--weighted regularizer on $\boldsymbol{\phi}$, which controls the trade-off between preserving pretrained behavior and adapting to the target task.}
  \label{fig:hip_mech}
\end{figure*}

\paragraph{Overview.}
HiP-LoRA represents each adapter update in the cached SVD basis of the pretrained weight and decomposes it into two channels:
\begin{equation}
\label{eq:hip_total_main}
\Delta\mathbf{W}
=
\underbrace{\mathbf{U}_k\operatorname{diag}(\boldsymbol{\phi})\mathbf{V}_k^\top}_{\text{principal channel}}
+
\underbrace{s\,\Delta\mathbf{W}_{\mathrm{res}}}_{\text{residual channel}}.
\end{equation}
The principal channel edits the dominant singular subspace, while the residual channel captures complementary low-rank adaptation in the two-sided orthogonal complement.
The key idea is to avoid two extremes: editing dominant singular directions without control, or excluding them altogether.
HiP-LoRA instead allows limited edits in the principal subspace while reserving the complementary subspace for residual adaptation.
To control the trade-off between preserving pretrained behavior and adapting to the target task, we regularize $\boldsymbol{\phi}$ with a singular-value--weighted quadratic budget.
Figure~\ref{fig:hip_mech} summarizes the two-channel design.

\paragraph{Principal channel: controlled editing in the dominant subspace.}
We parameterize the principal update as
\begin{equation}
\label{eq:principal_channel}
\Delta\mathbf{W}_{\parallel}
=
\mathbf{U}_k\operatorname{diag}(\boldsymbol{\phi})\mathbf{V}_k^\top,
\end{equation}
where $\boldsymbol{\phi}\in\mathbb{R}^k$ contains one learned gain deviation for each cached singular direction.
This parameterization is simple, interpretable, and restricted to the dominant singular directions.
Unlike hard exclusion, it still allows task-relevant adjustments in the principal subspace when exact preservation would be too restrictive.

\paragraph{Residual channel: complementary low-rank adaptation.}
To preserve adaptation capacity outside the dominant singular subspace, we apply a rank-$r$ LoRA update restricted to the two-sided orthogonal complement:
\begin{equation}
\label{eq:residual_channel_main}
\Delta\mathbf{W}_{\mathrm{res}}
=
(I-\mathbf{P}_U)\,(\mathbf{B}\mathbf{A})\,(I-\mathbf{P}_V).
\end{equation}
Equivalently, this can be realized by projecting the low-rank factors,
\begin{equation}
\label{eq:factor_proj}
\widetilde{\mathbf{B}}=(I-\mathbf{U}_k\mathbf{U}_k^\top)\mathbf{B},\qquad
\widetilde{\mathbf{A}}=\mathbf{A}(I-\mathbf{V}_k\mathbf{V}_k^\top),
\end{equation}
and using
\begin{equation}
\label{eq:res_proj}
\Delta\mathbf{W}_{\mathrm{res}}=\widetilde{\mathbf{B}}\widetilde{\mathbf{A}}.
\end{equation}
By construction,
\[
\mathbf{U}_k^\top\Delta\mathbf{W}_{\mathrm{res}}=\mathbf{0},
\qquad
\Delta\mathbf{W}_{\mathrm{res}}\mathbf{V}_k=\mathbf{0},
\]
so the residual channel does not reuse capacity assigned to the principal channel.
This gives the two channels distinct roles: controlled editing inside the dominant subspace and residual adaptation outside it.
In practice, we enforce the complement constraint by factor retraction once per optimizer step, which preserves the standard LoRA low-rank forward path; implementation details are deferred to Appendix~\ref{app:projection}.

\paragraph{Spectrum-aware stability control.}
Editing dominant pretrained directions is useful but potentially destabilizing.
We therefore regularize the principal deviations with a singular-value--weighted quadratic penalty:
\begin{equation}
\label{eq:hip_budget}
\Omega(\boldsymbol{\phi})
=
\sum_{i=1}^{k} w_i\,\phi_i^2,
\qquad
w_i=\frac{\sigma_i^\gamma}{\sum_{j=1}^{k}\sigma_j^\gamma}.
\end{equation}
Here $\gamma\ge 0$ controls how strongly larger-singular-value directions are protected.
When $\gamma=0$, all principal directions are penalized equally; larger $\gamma$ places more weight on high-$\sigma$ directions.
Normalizing the weights keeps the overall penalty scale comparable across layers and choices of $k$.

\paragraph{Objective.}
We initialize $\boldsymbol{\phi}=\mathbf{0}$ and the residual factors to zero, so that the effective weight matches the pretrained one at initialization.
Given task loss $\mathcal{L}_{\mathrm{task}}$, HiP-LoRA optimizes
\begin{equation}
\label{eq:hip_objective}
\min_{\boldsymbol{\phi},\mathbf{A},\mathbf{B}}
\ \mathcal{L}_{\mathrm{task}}(\mathbf{W}+\Delta\mathbf{W})
+\lambda_{\mathrm{stab}}\Omega(\boldsymbol{\phi}),
\end{equation}
where $\Delta\mathbf{W}$ is defined in Eq.~\ref{eq:hip_total_main}.
The hyperparameter $\lambda_{\mathrm{stab}}$ directly controls the stability--plasticity trade-off:
the principal channel allows limited edits in dominant directions, while the complementary residual channel preserves adaptation capacity outside them.
For implementation, an equivalent reparameterization $\boldsymbol{\theta}=\boldsymbol{\sigma}+\boldsymbol{\phi}$ may be used; we keep the main text in terms of $\boldsymbol{\phi}$ for clarity.

\section{Experiments and Analysis}
\label{sec:experiments}

We evaluate HiP-LoRA from three perspectives:
(1) \textbf{Main results:} We evaluate single-task adaptation and training-free multi-adapter merging to assess retention, downstream performance, and adapter compatibility.
(2) \textbf{Mechanistic analysis:} We examine stability--plasticity control through sweeps over $(\lambda_{\text{stab}},\gamma)$ and analyze the resulting spectral allocation patterns.
(3) \textbf{Additional evidence:} We include a diagnostic probe of spectrum-dependent sensitivity and report results in continual tuning and knowledge editing.
Ablations further isolate the contribution of each component.

\subsection{Experimental setup}
\label{subsec:exp_setup}

\paragraph{Benchmarks and backbone.}
We evaluate HiP-LoRA on four domains: general knowledge (\textsc{MMLU}), science QA (\textsc{ScienceQA}), mathematical reasoning (\textsc{GSM8K}), and code generation (CodeAlpaca-20k $\rightarrow$ \textsc{HumanEval}).
Unless noted otherwise, we use standard zero-shot prompting, and report Pass@1 on \textsc{HumanEval}.
For the \textsc{MMLU} domain adapter, we fine-tune on Alpaca-52k and evaluate on the MMLU test set.
Unless stated otherwise, all experiments use \textbf{Llama-3.1-8B}; dataset versions, prompts, and decoding details are given in Appendix~\ref{app:datasets}.

\paragraph{Protocols and metrics.}
Table~\ref{tab:single_task_main} reports downstream performance on \textsc{MMLU}, \textsc{ScienceQA}, \textsc{GSM8K}, and \textsc{HumanEval}.
Results are aggregated over four independently trained single-task adapters, one per domain.
To measure preservation, we report \textbf{Retain}, the mean out-of-domain score drop on a fixed retention suite (lower is better, in percentage points).
For an adapter trained on domain $d$, we exclude $d$ itself from the retention suite.
For composability, we merge independently trained adapters by standard weight-space addition, without joint optimization or post-merge retraining, and report \textbf{Before/After/$\Delta$} in Table~\ref{tab:merge_main}.
We also report \textbf{MergeFail}, the fraction of sampled merges whose post-merge score falls below a fixed fraction $\tau$ of the corresponding single-task score.
Full metric definitions, thresholds, and sampling details are deferred to Appendix~\ref{app:retain_def} and Appendix~\ref{app:merging}.

\paragraph{Baselines and hyperparameters.}
\label{sec:baselines}
We compare against \textbf{LoRA}~\cite{hu2022lora}, \textbf{PiSSA}~\cite{meng2024pissa}, \textbf{Split-LoRA}~\cite{lin2024splitlora} ($4{\times}4$), and \textbf{Proj-LoRA}, which restricts a scaled LoRA update to the two-sided complement of the top-$k$ SVD subspace,
\[
\Delta W=(I-U_kU_k^\top)\,(s\,BA)\,(I-V_kV_k^\top),
\]
with $s=\alpha/r$.
Unless stated otherwise, all methods adapt self-attention projection layers only (\texttt{q\_proj, k\_proj, v\_proj, o\_proj}) for fair comparison.
HiP-LoRA uses $k{=}32$ and residual rank $r{=}16$ by default, and sweeps $\lambda_{\text{stab}}\in\{0,0.01,0.03,0.1,0.3,1.0\}$ and $\gamma\in\{0,0.5,1.0,2.0\}$.
Split-LoRA is included mainly for merging and is omitted from Table~\ref{tab:single_task_main}.
All results are mean $\pm$ std over 3 seeds; implementation details, selection protocol, stronger baselines, and efficiency measurements are deferred to Appendix~\ref{app:impl_details}--\ref{app:extra_baselines} and Table~\ref{tab:efficiency}.

\subsection{Main results}
\subsubsection{Single-task adaptation}
\label{subsec:single_task}

As primary evidence for improved retention, Table~\ref{tab:single_task_main} shows a consistent pattern: HiP-LoRA improves or matches downstream task accuracy while sharply reducing pretraining degradation (Retain).
Relative to standard LoRA at the same residual rank ($r{=}16$), HiP-LoRA cuts Retain by more
than half, yet still improves performance across knowledge (\textsc{MMLU}), reasoning
(\textsc{GSM8K}), code (\textsc{HumanEval}), and science QA (\textsc{ScienceQA}).

This setting is particularly sensitive to interference: although updates are low-rank, their energy can concentrate on high-$\sigma$ directions that underpin broad pretrained competence, causing disproportionate retention degradation.
HiP-LoRA mitigates this by stabilizing principal-subspace edits via the $\sigma$-weighted budget while preserving task capacity through the projected residual channel.
As a result, it learns task-specific behavior with substantially less collateral damage to pretrained capabilities.
Notably, Proj-LoRA improves retention over LoRA but still lags behind HiP-LoRA, indicating that pure exclusion from principal directions is not enough; controlled, spectrum-aware modulation within the principal subspace is important for the retention--accuracy balance. HiP-LoRA also outperforms AdaLoRA under matched budgets (Appendix Table~\ref{tab:adalora_appendix}), suggesting that the improvement is not explained by adaptive rank allocation alone.

\begin{table*}[t]
  \centering
  \small
  \setlength{\tabcolsep}{6pt}
  \renewcommand{\arraystretch}{1.10}
  \begin{threeparttable}
  \caption{\textbf{Single-task adaptation on Llama-3.1-8B} (mean $\pm$ std over 3 seeds).}
  \label{tab:single_task_main}
  \begin{tabular}{lccccc}
    \toprule
    \textbf{Method} &
    \textbf{MMLU} $\uparrow$ &
    \textbf{GSM8K} $\uparrow$ &
    \textbf{HumanEval(Pass@1)} $\uparrow$ &
    \textbf{ScienceQA} $\uparrow$ &
    \textbf{Retain (pp)} $\downarrow$ \\
    \midrule
    LoRA ($r$=16)                     & $37.33\!\pm\!0.75$ & $55.64\!\pm\!0.25$ & $29.54\!\pm\!0.80$ & $92.26\!\pm\!0.35$ & $12.3\!\pm\!1.1$ \\
    Proj-LoRA ($r$=16)                & $37.41\!\pm\!0.49$ & $55.12\!\pm\!0.38$ & $29.10\!\pm\!0.64$ & $92.08\!\pm\!0.51$ &  $8.9\!\pm\!0.9$ \\
    PiSSA ($r$=16)            & $38.68\!\pm\!0.61$ & $56.63\!\pm\!0.36$ & $30.92\!\pm\!0.70$ & $93.62\!\pm\!0.33$ &  $7.1\!\pm\!0.7$ \\
    \textbf{HiP-LoRA ($k$=32, $r$=16)} & $\mathbf{39.44\!\pm\!0.58}$ & $\mathbf{57.36\!\pm\!0.34}$ & $\mathbf{31.52\!\pm\!0.66}$ & $\mathbf{94.05\!\pm\!0.29}$ & $\mathbf{4.9\!\pm\!0.6}$ \\
    \bottomrule
  \end{tabular}
  \end{threeparttable}
  \vspace{-1.0ex}
\end{table*}

\subsubsection{Training-free multi-adapter merging}
\label{subsec:merging}

As our primary test of adapter compatibility, merging independently trained adapters constitutes a worst-case interference setting:
even if each single-task adapter appears benign in isolation, additive composition can push
the model far from the pretrained operating point.
This setting isolates structural compatibility of updates, independent of optimization
dynamics.

Table~\ref{tab:merge_main} highlights this failure mode for standard LoRA.
After merging, performance degradation is highly non-uniform and often collapses on specific
domains (e.g., \textsc{ScienceQA}), consistent with the accumulation of drift along spectrally
sensitive directions.

Under the same training-free merge operator, HiP-LoRA is substantially more robust.
Across all reported metrics, it reduces the mean absolute post-merge drop from 20.24 to 5.42
points (a 73.2\% relative reduction versus LoRA, $r{=}16$), while maintaining strong
single-task performance.
This is consistent with our central claim: by explicitly budgeting principal-subspace edits and isolating
remaining capacity in the orthogonal residual channel, HiP-LoRA produces adapters whose updates
are more additively compatible.

We report MergeFail (threshold $\tau{=}0.9$) and full merge-sampling details in Appendix~\ref{app:merging}.
The same trend holds under a stronger merge rule. With TIES-Merging applied to the same trained adapters, HiP-LoRA reduces the mean absolute post-merge drop from 13.8 to 4.7 points and MergeFail from 38\% to 4.5\% relative to LoRA (Appendix Table~\ref{tab:ties_appendix}).
Because all adapters are trained independently and merged without joint optimization, this improvement reflects better structural compatibility of the learned updates rather than a favorable training procedure.

\begin{table*}[t]
\centering
\small
\setlength{\tabcolsep}{3.2pt}
\renewcommand{\arraystretch}{1.10}

\begin{threeparttable}
\caption{\textbf{Multi-task performance before/after training-free merging on Llama-3.1-8B.}
We evaluate LoRA variants on MMLU, ScienceQA, GSM8K (\textbf{accuracy}) and HumanEval (\textbf{Pass@\{1,5,10\}}).
\textbf{Before}: single-task adapter performance. \textbf{After}: performance after merging. $\Delta$ is (After$-$Before).
We report results for merging all four single-task adapters (merge size $t{=}4$).
When we report the ``mean absolute post-merge drop'' in the text, we average $|\Delta|$ over the six reported metrics.}

\label{tab:merge_main}

\begin{tabular*}{\textwidth}{@{\extracolsep{\fill}}llc ccc ccc@{}}
\toprule
\textbf{Model} & \textbf{Method} & \makecell{\textbf{Merge}\\\textbf{Status}} &
\textbf{MMLU} & \textbf{ScienceQA} & \textbf{GSM8K} &
\multicolumn{3}{c}{\textbf{HumanEval}} \\
\cmidrule(lr){7-9}
& & & & & & \textbf{Pass@1} & \textbf{Pass@5} & \textbf{Pass@10} \\
\midrule

\multirow{15}{*}{\makecell{Llama\\3.1-8B}}
& \multirow{3}{*}{LoRA($r$=16)}
& Before & $37.33\!\pm\!0.75$ & $92.26\!\pm\!0.35$ & $55.64\!\pm\!0.25$ & $29.54\!\pm\!0.80$ & $52.98\!\pm\!0.70$ & $62.45\!\pm\!0.80$ \\
& & After  & $31.37\!\pm\!1.10$ & $32.15\!\pm\!1.90$ & $25.05\!\pm\!1.40$ & $17.04\!\pm\!2.20$ & $45.72\!\pm\!1.60$ & $57.42\!\pm\!1.50$ \\
& & $\Delta$ (pp) & $-5.96$ & $-60.11$ & $-30.59$ & $-12.50$ & $-7.26$ & $-5.03$ \\
\cmidrule(lr){2-9}

& \multirow{3}{*}{Split-LoRA($4{\times}4$)}
& Before & $37.20\!\pm\!0.70$ & $92.00\!\pm\!0.45$ & $55.55\!\pm\!0.30$ & $29.90\!\pm\!0.90$ & $53.00\!\pm\!0.80$ & $62.40\!\pm\!0.85$ \\
& & After  & $31.29\!\pm\!1.05$ & $35.89\!\pm\!1.45$ & $26.95\!\pm\!1.15$ & $18.60\!\pm\!1.70$ & $45.64\!\pm\!1.30$ & $57.62\!\pm\!1.20$ \\
& & $\Delta$ (pp) & $-5.91$ & $-56.11$ & $-28.60$ & $-11.30$ & $-7.36$ & $-4.78$ \\
\cmidrule(lr){2-9}

& \multirow{3}{*}{Proj-LoRA($r$=16) }
& Before  & $37.41\!\pm\!0.49$ & $92.08\!\pm\!0.51$ & $55.12\!\pm\!0.38$ & $29.10\!\pm\!0.64$ & $52.10\!\pm\!1.05$ & $61.22\!\pm\!0.88$ \\
& & After   & $33.92\!\pm\!0.91$ & $62.84\!\pm\!2.10$ & $41.78\!\pm\!1.62$ & $23.96\!\pm\!1.74$ & $49.18\!\pm\!1.44$ & $60.02\!\pm\!1.12$ \\
& & $\Delta$ (pp) & $-3.49$ & $-29.24$ & $-13.34$ & $-5.14$ & $-2.92$ & $\mathbf{-1.20}$ \\
\cmidrule(lr){2-9}

& \multirow{3}{*}{PiSSA ($r$=16)}
& Before  & $38.68\!\pm\!0.61$ & $93.62\!\pm\!0.33$ & $56.63\!\pm\!0.36$ & $30.92\!\pm\!0.70$ & $53.98\!\pm\!0.96$ & $63.52\!\pm\!0.73$ \\
& & After   & $35.41\!\pm\!0.88$ & $71.18\!\pm\!1.84$ & $45.66\!\pm\!1.41$ & $25.92\!\pm\!1.53$ & $50.86\!\pm\!1.22$ & $61.84\!\pm\!0.98$ \\
& & $\Delta$ (pp) & $-3.27$ & $-22.44$ & $-10.97$ & $-5.00$ & $-3.12$ & $-1.68$ \\
\cmidrule(lr){2-9}

& \multirow{3}{*}{HiP-LoRA ($k$=32, $r$=16)}
& Before  & $39.44\!\pm\!0.58$ & $94.05\!\pm\!0.29$ & $57.36\!\pm\!0.34$ & $31.52\!\pm\!0.66$ & $55.12\!\pm\!0.91$ & $64.88\!\pm\!0.79$ \\
& & After   & $37.02\!\pm\!0.76$ & $79.06\!\pm\!1.52$ & $49.64\!\pm\!1.18$ & $27.78\!\pm\!1.27$ & $52.92\!\pm\!1.03$ & $63.44\!\pm\!0.86$ \\
\rowcolor{gray!12}
& & $\Delta$ (pp) & $\mathbf{-2.42}$ & $\mathbf{-14.99}$ & $\mathbf{-7.72}$ & $\mathbf{-3.74}$ & $\mathbf{-2.20}$ & $-1.44$ \\
\bottomrule
\end{tabular*}

\end{threeparttable}
\end{table*}

\subsection{Controllable stability--plasticity trade-off and spectral allocation}
\label{subsec:c2_controllability}

\begin{figure}[t]
  \centering
  \includegraphics[width=0.9\linewidth]{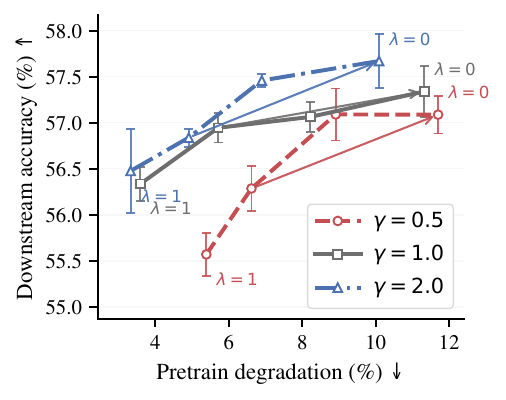}
\caption{\textbf{Controllable stability--plasticity trade-off.}
Downstream accuracy versus pretraining degradation (lower is better) under different values of $\lambda_{\text{stab}}$ and $\gamma$.
Varying $\lambda_{\text{stab}}$ traces a continuous trade-off curve, while $\gamma$ controls how strongly updates are biased away from large-$\sigma$ directions.
Points show means over 3 seeds; error bars indicate $\pm 1$ std.}
\label{fig:stability_plasticity}
\end{figure}

\begin{figure}[t]
  \centering
  \includegraphics[width=0.9\linewidth]{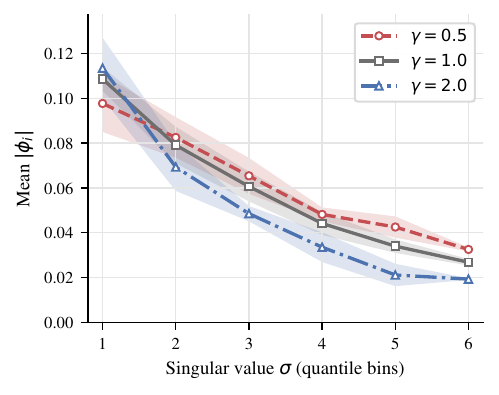}
  \caption{\textbf{Spectral allocation induced by HiP-LoRA.}
Binned averages of $|\phi_i|$ versus singular value $\sigma_i$ across layers.
HiP-LoRA assigns smaller updates to larger-$\sigma$ directions, with the strength of this bias controlled by $\gamma$.
Curves show means over 3 seeds; shaded bands indicate $\pm 1$ std.}
  \label{fig:spectrum_heatmap}
\end{figure}

To explain the main gains and validate controllability, HiP-LoRA exposes stability--plasticity trade-offs as an explicit, continuous control via $(\lambda_{\text{stab}},\gamma)$ rather than an implicit side effect of optimization.
The stability weight $\lambda_{\text{stab}}$ directly modulates the magnitude of principal-subspace edits,
while the orthogonal residual channel preserves task-specific adaptation capacity.
As shown in Figure~\ref{fig:stability_plasticity}, varying $\lambda_{\text{stab}}$ traces a smooth trade-off curve
between downstream performance and pretraining retention. Concretely, increasing $\lambda_{\text{stab}}$ suppresses the energy of principal-subspace edits,
while leaving residual capacity available for task learning.
The spectral sensitivity parameter $\gamma$ further adjusts how strongly large-$\sigma$ directions are protected,
yielding a family of controllable operating regimes. To verify that HiP-LoRA enforces its intended spectral allocation,
we inspect the learned principal deviations $\phi$ in SVD coordinates.
Figure~\ref{fig:spectrum_heatmap} shows a graded, spectrum-aligned allocation in which larger singular values receive smaller edits, consistent with the singular-value--weighted budget and indicating direction-level control rather than undifferentiated shrinkage.

\subsection{Spectral importance sanity check}
\label{subsec:c1_spectral_sensitivity}

As a diagnostic motivation probe (not a causal proof), we examine whether functional sensitivity is spectrum-dependent in the SVD geometry of pretrained weights.
This probe is intended to justify protecting large-$\sigma$ directions more strongly, rather than to serve as the main evidence for our end-to-end gains.

\paragraph{Question.}
Do large singular values $\sigma_i$ actually correspond to functionally sensitive directions?
If so, a spectrum-aware stability budget is targeting the directions where small geometric drift is most likely to harm general competence.

\paragraph{Setup.}
On representative self-attention projection matrices, we compute a truncated SVD ($k{=}32$) and apply controlled rank-1 interventions along individual singular directions (Zero / Flip / Noise; details in Appendix~\ref{app:interventions}).
We then measure the induced performance \emph{drop} under fixed evaluation conditions.

\paragraph{What we see.}
The most pronounced effect is observed for \textbf{Zero} (removing a direction): performance drops show a strong positive correlation with $\sigma$ (Table~\ref{tab:corr}), suggesting that large-$\sigma$ directions can be disproportionately sensitive under destructive interventions.
In contrast, \textbf{Flip} and small \textbf{Noise} yield weaker effects, consistent with a picture where \emph{catastrophic} perturbations expose sensitivity more reliably than mild local distortions. We emphasize effect sizes with confidence intervals in the main text; for completeness, we additionally report p-values/significance tests in the appendix, since this experiment is meant as a sanity-check probe.

\paragraph{Why this matters for HiP-LoRA.}
This sanity check motivates our design choice to protect large-$\sigma$ directions \emph{more} than small-$\sigma$ ones: the stability regularizer in HiP-LoRA is designed to impose this bias, while still allowing plasticity to flow through the residual orthogonal channel.
We emphasize the limits of the evidence (layer dependence; weaker signals for Flip/Noise) and provide full visualizations in Appendix~\ref{app:figs}.

\begin{table}[t]
\centering
\small
\caption{Sanity check: correlation between singular value magnitude and performance drop under rank-1 interventions. We report Pearson $r$ computed on $\log(1+\sigma)$ with bootstrap 95\% CI.}
\label{tab:corr}
\begin{tabular}{lccc}
\toprule
Perturb & $n$ & $r$ & 95\% CI \\
\midrule
flip       & 20  & 0.33 & [$-0.26$, 0.75] \\
noise\_0.1 & 127 & 0.14 & [$-0.07$, 0.34] \\
zero       & 29  & \textbf{0.86} & \textbf{[0.14, 0.95]} \\
\bottomrule
\end{tabular}
\end{table}

\subsection{Continual instruction tuning and knowledge editing}
\label{subsec:continual_and_editing}

Beyond single-task adaptation and training-free merging, we further test HiP-LoRA in two interference-sensitive regimes: continual instruction tuning and knowledge editing (full protocols and tables in Appendix~\ref{app:continual_edit}).
Both settings require localized updates: if adaptation spills into spectrally dominant directions, it can lead to forgetting in continual learning or poor locality in editing.

\textbf{Continual instruction tuning.}
Appendix Table~\ref{tab:continual_brief} shows that HiP-LoRA achieves the best final AvgAcc and the lowest forgetting under matched budgets, improving AvgAcc to 74.6 while reducing forgetting to 6.9, compared with 73.1/10.2 for PiSSA and 71.8/13.6 for LoRA.

\textbf{Knowledge editing.}
Appendix Table~\ref{tab:editing_brief} shows that HiP-LoRA achieves the highest edit success with the smallest locality change, reaching 89.2 edit success with 2.1 pp locality change, compared with 87.9/3.1 for PiSSA and 86.4/3.8 for LoRA.

\subsection{Ablations and efficiency}
\label{subsec:ablations}
\label{subsec:complexity}

\begin{figure}[t]
  \centering
  \includegraphics[width=1.0\linewidth]{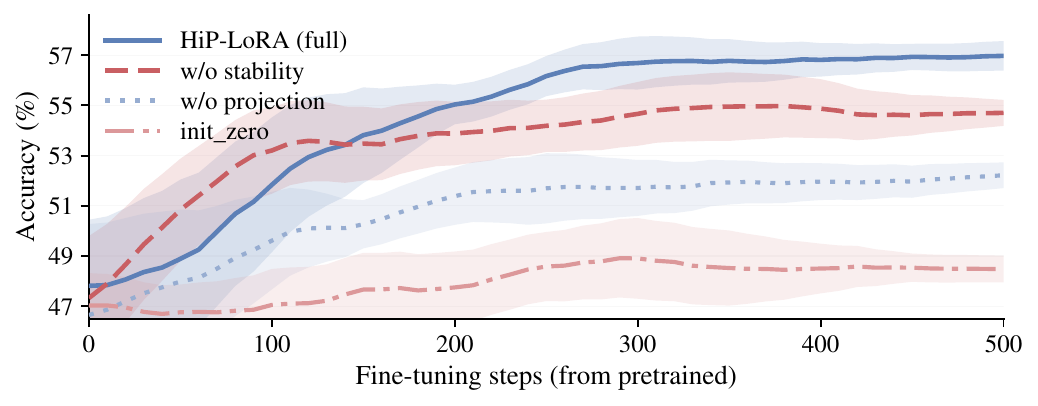}
  \caption{\textbf{Optimization dynamics under ablations.}
Downstream accuracy versus fine-tuning steps from the pretrained initialization.
Removing the stability budget, orthogonal projection, or pretrained-aligned initialization leads to degraded or less stable optimization trajectories. Curves are mean over 3 seeds; shaded regions denote $\pm$1 std.}
  \label{fig:ablation_dynamics}
\end{figure}

\paragraph{Ablations.}
Figure~\ref{fig:ablation_dynamics} analyzes optimization dynamics under controlled ablations, isolating which components are necessary for the main retention and merging gains.
Removing the spectrum-aware stability budget or the orthogonal residual projection leads to slower or less stable convergence, even when final accuracy differences are modest.
Appendix Table~\ref{tab:ablation_summary} shows the same pattern numerically: removing the stability regularizer raises Retain from 4.9 to 9.8 and MergeFail from 6.8 to 14.2, while removing the residual projection raises MergeFail to 12.5.
These results support the two-channel design: stabilization alone is not enough without isolating residual capacity in the two-sided complement.

\paragraph{Efficiency.}
HiP-LoRA adds only modest overhead in return.
It requires a one-time top-$k$ randomized SVD for each adapted frozen weight matrix, which is cached and reused across training and merging.
For each adapted matrix $W\in\mathbb{R}^{m\times n}$, storing $(U_k,V_k,\sigma)$ adds $k(m+n)+k$ scalars.

In our Llama-3.1-8B setting, throughput decreases from 42k tok/s for LoRA to 40k tok/s for HiP-LoRA, peak memory changes from 78.0 GB to 78.6 GB, and the one-time SVD preprocessing takes 18 minutes with a 0.13 GB cache (Appendix Table~\ref{tab:efficiency}).
The residual path remains in the same LoRA-style low-rank form and does not materialize dense $BA$ during training or adapter-mode inference.

Training-free merging also retains the same asymptotic form as LoRA-style merging: adapters are combined by adding their realized updates in weight space.
Our ``no dense $BA$'' claim applies to training and adapter-mode computation; when an adapter is folded into the backbone for deployment, the dense $\Delta W$ may be materialized once per module, as in standard LoRA merging, so the deployed runtime footprint remains unchanged.

\section{Discussion and Limitations}
\label{sec:limitations}
\label{subsec:tradeoffs}

HiP-LoRA trades expressiveness for control. The principal channel uses direction-wise gain editing on the top-$k$ singular directions, while the residual channel is restricted to the two-sided orthogonal complement. This keeps the update simple and stable, but it can be restrictive when a task requires larger changes along dominant directions. In such cases, the method can be relaxed by decreasing $\lambda_{\text{stab}}$, reducing the spectral bias through smaller $\gamma$, or increasing $k$.

This design introduces a one-time SVD cost for frozen weight matrices. Although the factors can be cached and reused, the overhead may still be non-trivial for very large models or when many modules are adapted. More expressive principal-channel parameterizations, or residual updates that retain cross-subspace interactions, may further improve capacity at the cost of weaker control.

We focus on LLM adaptation in this work. The same idea may be relevant to other models built around large frozen linear projections, but that remains to be tested. A broader evaluation is left to future work.
\section{Conclusion}
We presented HiP-LoRA, an SVD-based PEFT framework that allocates LoRA updates into two complementary channels: a constrained edit in the top-$k$ principal singular subspace and a plastic residual update restricted to the orthogonal complement.
By introducing a singular-value-aware, direction-wise stability budget on the principal channel, HiP-LoRA provides continuously tunable control over how much dominant pre-trained directions are edited, bridging hard protection with unconstrained adaptation.
Across continual instruction tuning, adapter merging, and knowledge editing, HiP-LoRA improves retention, merge stability, and edit locality over LoRA and representative subspace baselines under comparable parameter budgets.
Moreover, analyses in the SVD coordinate system reveal a graded pattern of learned update magnitudes aligned with the singular spectrum, offering an interpretable view of the stability--plasticity trade-off in PEFT.

\bibliographystyle{named}
\bibliography{ijcai26}

\FloatBarrier        
\clearpage  
\appendix

\section{Theoretical Derivations and Method Details}
\label{app:theory_derivations}

\label{sec:appendix}

\subsection{Exact update forms}
\label{app:exact_forms}

For each adapted \emph{frozen} backbone matrix $W_0\in\mathbb{R}^{m\times n}$, we precompute and cache a truncated SVD
\[
W_0\approx U_k\operatorname{diag}(\sigma)V_k^\top,
\]
with $U_k\in\mathbb{R}^{m\times k}$, $V_k\in\mathbb{R}^{n\times k}$, and $\sigma\in\mathbb{R}_{+}^k$.
We also define the cached residual
\[
\widetilde W := W_0 - U_k\operatorname{diag}(\sigma)V_k^\top.
\]

Consistent with the main text, we write the HiP-LoRA update as
\begin{equation}
\label{eq:hip_total_appendix}
\Delta W
=
\underbrace{U_k\operatorname{diag}(\phi)V_k^\top}_{\Delta W_{\parallel}}
+
\Delta W_{\mathrm{res}},
\end{equation}
where $\phi\in\mathbb{R}^k$ denotes the learned principal deviations and $\Delta W_{\mathrm{res}}$ is the residual-channel update in the two-sided orthogonal complement.
The corresponding effective weight is therefore
\[
W_{\text{eff}} = W_0 + \Delta W
= \widetilde W + U_k\operatorname{diag}(\sigma+\phi)V_k^\top + \Delta W_{\mathrm{res}}.
\]

In implementation, one may equivalently introduce
\[
\theta := \sigma + \phi,
\]
so that
\[
W_{\text{eff}} = \widetilde W + U_k\operatorname{diag}(\theta)V_k^\top + \Delta W_{\mathrm{res}}.
\]
We keep the appendix aligned with the main text and use $\phi$ as the primary variable; the $\theta$ parameterization is only an equivalent implementation form.

\subsection{Efficient projection}
\label{app:projection}

We implement the two-sided complement projection without forming dense matrices.
Define the principal-subspace projectors
$P_U := U_kU_k^\top \in \mathbb{R}^{m\times m}$ and
$P_V := V_kV_k^\top \in \mathbb{R}^{n\times n}$.
With LoRA factors $B\in\mathbb{R}^{m\times r}$ and $A\in\mathbb{R}^{r\times n}$, we apply the exact factor retraction
$\widetilde B=(I-P_U)B$ and $\widetilde A=A(I-P_V)$, and use $\Delta W_{\mathrm{res}}=\widetilde B\,\widetilde A$
(Eq.~\ref{eq:factor_proj}--\ref{eq:res_proj}).
This keeps the forward pass in the standard LoRA form (two GEMMs) and avoids materializing the dense matrix $BA$.
For conceptual reference, letting $X:=BA$, the equivalent projector form satisfies
\[
\begin{aligned}
(I-P_U)X(I-P_V)
  &= X-U_k(U_k^\top X)-(X V_k)V_k^\top \\
  &\quad +U_k(U_k^\top X V_k)V_k^\top,
\end{aligned}
\]
and each term can be evaluated via low-rank multiplications (e.g., $U_k^\top X=(U_k^\top B)A$ and $X V_k=B(A V_k)$),
so neither $X$ nor the dense projectors are formed in practice.

\subsection{Spectrum-aware stability regularization}
\label{app:stability}

We penalize principal deviations with a singular-value--weighted quadratic budget:
\begin{equation}
\Omega(\phi)=\sum_{i=1}^{k} w_i\,\phi_i^2,\qquad
w_i=\frac{\sigma_i^\gamma}{\sum_{j=1}^k\sigma_j^\gamma},\qquad
\sum_{i=1}^k w_i = 1.
\end{equation}
The stability weight is $\lambda_{\text{stab}}$, and $\gamma$ controls how strongly larger-singular-value directions are protected.
Normalizing the weights keeps the overall penalty scale comparable across layers and choices of $k$.

For a task dataset $\mathcal{D}$, the full objective is
\begin{equation}
\label{eq:hip_objective_recap_appendix}
\min_{\phi,A,B}\;
\mathcal{L}\!\big(\mathcal{D};\,W_0 + U_k\operatorname{diag}(\phi)V_k^\top + \Delta W_{\mathrm{res}}(A,B)\big)
+\lambda_{\text{stab}}\,\Omega(\phi).
\end{equation}

If the implementation uses the equivalent reparameterization $\theta=\sigma+\phi$, the same objective can be written in terms of $\theta$, but we keep the appendix consistent with the main text and optimize with respect to $\phi$ in the presentation.

\subsection{Training algorithm}
\label{app:algorithm}

\begin{algorithm}[t]
\caption{HiP-LoRA (recap).}
\label{alg:hiplora_recap}
\begin{algorithmic}[1]
\STATE For each adapted matrix, precompute and cache top-$k$ SVD triplets $(U_k,V_k,\sigma)$ of the frozen backbone weight $W$.
\STATE Initialize $\phi\leftarrow 0$, $A\leftarrow 0$, $B\leftarrow 0$, and $\widetilde W\leftarrow W-U_k\operatorname{diag}(\sigma)V_k^\top$.
\FOR{training steps}
\STATE Project factors: $\widetilde B\leftarrow (I-P_U)B$, $\widetilde A\leftarrow A(I-P_V)$.
\STATE Form the update
\[
\Delta W = U_k\operatorname{diag}(\phi)V_k^\top + s\,\widetilde B\,\widetilde A,
\]
with the standard LoRA scaling $s=\alpha/r$ applied only to the residual channel.
\STATE Update $(\phi,A,B)$ by AdamW on task loss $+\lambda_{\text{stab}}\Omega(\phi)$ (adapter parameters only).
\STATE Retract $(B,A)\leftarrow(\widetilde B,\widetilde A)$ to enforce the two-sided complement constraint.
\ENDFOR
\end{algorithmic}
\end{algorithm}

\section{Implementation Details and Reproducibility}
\label{app:impl_details}

\subsection{Implementation notes}
\label{app:impl_notes}

\paragraph{Two-channel parameterization.}
We implement HiP-LoRA by extending standard PEFT LoRA modules with
(i) an additional principal-channel parameter vector $\phi\in\mathbb{R}^k$, and
(ii) cached SVD factors $(U_k,V_k,\sigma_{1:k})$ for each adapted matrix from the frozen backbone weight.

For each adapted weight matrix $W\in\mathbb{R}^{m\times n}$, we precompute a truncated SVD of the frozen weight:
\[
W \approx U_k \operatorname{diag}(\sigma_{1:k}) V_k^\top,
\qquad
U_k\in\mathbb{R}^{m\times k},\; V_k\in\mathbb{R}^{n\times k}.
\]
We cache
\[
\widetilde W = W - U_k \operatorname{diag}(\sigma_{1:k}) V_k^\top,
\]
and write the effective weight as
\[
W_{\text{eff}}
=
W + \Delta W
=
W + U_k \operatorname{diag}(\phi) V_k^\top + \Delta W_{\mathrm{res}}.
\]
Equivalently,
\[
W_{\text{eff}}
=
\widetilde W + U_k\operatorname{diag}(\sigma+\phi)V_k^\top + \Delta W_{\mathrm{res}}.
\]

In implementation, one may store $\theta=\sigma+\phi$ instead of $\phi$ directly.
This is only an equivalent reparameterization; throughout the paper we describe the method in terms of the principal deviations $\phi$.

The residual channel is the two-sided complement projection of a low-rank LoRA update.
In practice, we realize it as $\Delta W_{\mathrm{res}}=\widetilde B\,\widetilde A$ via factor projection, so the residual path remains in the standard LoRA form and no dense $BA$ is formed.

\paragraph{Target modules.}
Unless otherwise stated, we apply HiP-LoRA (and all LoRA-family baselines) to self-attention projection layers only (\texttt{q\_proj, k\_proj, v\_proj, o\_proj}) for all experiments, matching Sec.~\ref{sec:method} for fair budget comparison.

\paragraph{Efficient two-sided projection for the residual channel.}
Let $X=BA$ denote the (conceptual) low-rank update with $B\in\mathbb{R}^{m\times r}$ and $A\in\mathbb{R}^{r\times n}$.
Define the principal-subspace projectors $P_U:=U_kU_k^\top$ and $P_V:=V_kV_k^\top$.
The residual-channel update is the two-sided complement projection
\[
\Delta W_{\mathrm{res}}
=(I-P_U)\,(BA)\,(I-P_V)
=(I-U_kU_k^\top)\,(BA)\,(I-V_kV_k^\top).
\]
In practice, we implement this projection on the low-rank factors and never materialize $BA$:
\[
\widetilde B=(I-U_kU_k^\top)B,\qquad
\widetilde A=A(I-V_kV_k^\top),\qquad
\Delta W_{\mathrm{res}}=\widetilde B\,\widetilde A.
\]
The residual update is then applied in the standard LoRA form (two GEMMs), while the projection is enforced by a post-update retraction once per optimizer step.

A reference implementation is:
\begin{Verbatim}[breaklines,breakanywhere,fontsize=\small]
# U: (m,k), V: (n,k), A: (r,n), B: (m,r)
# Project factors (no dense BA is formed)
# U.T @ B: (k,r) and (A @ V): (r,k)
B_tilde = B - U @ (U.T @ B)      # (I - U U^T) B
A_tilde = A - (A @ V) @ V.T      # A (I - V V^T)
# Use DeltaW_res implicitly as (B_tilde @ A_tilde) in standard LoRA forward
\end{Verbatim}

\subsection{Training hyperparameters and budget matching}
\label{app:training_hparams}
\label{app:hyperparams}

Unless otherwise stated, all runs use AdamW
($\beta_1{=}0.9,\beta_2{=}0.95,\epsilon{=}1\mathrm{e}{-}8$), learning rate $2\mathrm{e}{-}4$
(3\% linear warmup, then constant), weight decay 0.01 on adapter parameters only, gradient clipping 1.0, sequence length 2{,}048 with packing, and a fixed budget of 2{,}000 optimizer steps per task.
All LoRA-family methods use dropout $=0.05$, scaling $\alpha{=}2r$, and the same self-attention projection targets.
We match trainable parameters within $\pm1\%$ and report mean $\pm$ std over three seeds.
Table~\ref{tab:appendix_hparams} summarizes the shared training, evaluation, and reproducibility settings.

\subsection{HiP-LoRA hyperparameters and selection protocol}
\label{app:hiplora_selection}
\label{app:hiplora_hparams}

\paragraph{HiP-LoRA hyperparameters.}
HiP-LoRA introduces method-specific hyperparameters controlling the spectrum-conditioned allocation of update energy.
Unless stated otherwise, we use principal size $k{=}32$ and residual rank $r{=}16$ for Llama-3.1-8B,
and report mean $\pm$ std over three seeds \texttt{[42,100,2024]}.

\paragraph{Choosing the principal size $k$.}
The principal size $k$ trades off stability and capacity.
If $k$ is too small, high-gain directions may be forced into the residual channel; if $k$ is too large,
the principal channel can dominate under non-trivial stability regularization and reduce task adaptivity.
We therefore sweep $k\in\{8,16,32,64\}$ and use $k{=}32$ by default.

\paragraph{Choosing the residual rank $r$ (capacity control).}
The residual rank $r$ plays a role analogous to the LoRA rank and primarily controls adaptation capacity in the orthogonal complement.
We sweep $r\in\{8,16,32\}$ under matched parameter budgets and use $r{=}16$ by default.

\paragraph{Stability weight $\lambda_{\text{stab}}$ (stability--plasticity control).}
We sweep $\lambda_{\text{stab}}\in\{0, 0.01, 0.03, 0.1, 0.3, 1.0\}$ (log-grid) on a held-out validation split and
select a balanced point based on downstream score and retention.
For readability, main trade-off plots use three representative settings $\lambda_{\text{stab}}\in\{0,0.03,0.3\}$.

\paragraph{Spectral sensitivity $\gamma$ (testing whether spectral weighting matters).}
We evaluate
\[
\gamma \in \{0,\ 0.5,\ 1,\ 2\},
\]
where $\gamma{=}0$ is an equal-weight control (non-spectral) to test whether improvements require spectrum-aware weighting.
We set $\gamma{=}1$ as the default.

\paragraph{MergeFail threshold $\tau$ and sensitivity.}
For training-free multi-adapter merging, we report MergeFail with the default threshold $\tau{=}0.9$ and include
a sensitivity check with $\tau\in\{0.8,0.9,0.95\}$.
Under simple addition, LoRA exhibits 62\% MergeFail whereas HiP-LoRA reduces it to 6.8\%
(Appendix Table~\ref{tab:ties_appendix}).

\paragraph{Randomized SVD details and verification.}
We compute top-$k$ factors using randomized SVD with power iterations $n_{\text{iter}}{=}2$ and stabilizer $\epsilon{=}10^{-6}$,
cache $(U_k,V_k,\sigma_{1:k})$ per layer in fp32 to avoid accumulation error, and reuse the cache across runs.
When oversampling is enabled, we use $p\in\{8,16\}$ (default $p{=}8$).
As a lightweight sanity check, we spot-check a small subset of layers against exact SVD to confirm comparable reconstruction error.

\paragraph{Compute-efficient selection protocol (coarse-to-fine).}
To minimize tuning compute while maintaining fairness, we adopt the following selection protocol:
(i) fix $(k,r,\gamma)=(32,16,1)$ and sweep $\lambda_{\text{stab}}$ on a held-out validation split to select a balanced point;
(ii) fix $(k,r,\lambda_{\text{stab}})$ at the selected balanced point and sweep $\gamma\in\{0,0.5,1,2\}$ to test whether spectral weighting is necessary;
(iii) fix $(r,\lambda_{\text{stab}},\gamma)$ at the selected values and sweep $k\in\{8,16,32,64\}$ to verify robustness to the principal size.
All baselines are tuned under the same learning-rate grid, step budget, target modules, and random seeds.

Unless otherwise stated, we visualize the stability--plasticity trade-off using three representative settings $\lambda_{\text{stab}}\in\{0,0.03,0.3\}$ for readability; the full log-grid and selection protocol are provided in Appendix~\ref{app:hiplora_hparams}.

\begin{table*}[t]
  \centering
  \small
  \setlength{\tabcolsep}{6pt}
  \renewcommand{\arraystretch}{1.08}
  \begin{threeparttable}
  \caption{\textbf{Training, evaluation, and reproducibility settings used in all experiments.}
  Unless stated otherwise, these settings are shared across all single-task runs, merging, continual tuning, and editing.}
  \label{tab:appendix_hparams}
  \begin{tabular}{llp{7.3cm}}
    \toprule
    \textbf{Category} & \textbf{Hyperparameter} & \textbf{Value / Notes} \\
    \midrule

    \multicolumn{3}{l}{\textbf{Backbone \& prompts}} \\
    \midrule
    & Backbone & Llama-3.1-8B (public release) \\
    & Tokenizer & Same as backbone \\
    & Prompt format & Alpaca-style: \texttt{\{instruction, input, response\}} \\
    & Max sequence length & 2048 tokens; packing enabled \\

    \addlinespace
    \multicolumn{3}{l}{\textbf{Optimization}} \\
    \midrule
    & Optimizer & AdamW ($\beta_1{=}0.9$, $\beta_2{=}0.95$, $\epsilon{=}1\mathrm{e}{-}8$) \\
    & Learning rate & $2\mathrm{e}{-}4$ (linear warmup for 3\% steps, then constant) \\
    & Weight decay & 0.01 on adapter parameters only \\
    & Gradient clipping & Global grad norm 1.0 \\
    & Precision & bf16 training \\

    \addlinespace
    \multicolumn{3}{l}{\textbf{Batching \& budget}} \\
    \midrule
    & Effective global batch size & 128 sequences / optimizer step (via gradient accumulation when needed) \\
    & Max updates & 2000 optimizer steps per task (compute-matched across methods) \\
    & Early stopping & None (fixed-step budget) \\

    \addlinespace
    \multicolumn{3}{l}{\textbf{LoRA-family baselines}} \\
    \midrule
    & LoRA rank $r_{\text{LoRA}}$ & $\{8,16,32\}$; we report the tuned choice per setting \\
    & LoRA scaling $\alpha$ & $\alpha=2r_{\text{LoRA}}$ \\
    & LoRA dropout & 0.05 \\
    & Target modules & Self-attention projections only: \texttt{q\_proj, k\_proj, v\_proj, o\_proj} \\

    \addlinespace
    \multicolumn{3}{l}{\textbf{HiP-LoRA (method-specific)}} \\
    \midrule
    & Principal size $k$ & 32 \\
    & Residual rank $r$ & 16 \\
    & Stability weight $\lambda_{\text{stab}}$ & $\{0, 0.01, 0.03, 0.1, 0.3, 1.0\}$ (log-grid) \\
    & Spectral sensitivity $\gamma$ & $\{0, 0.5, 1, 2\}$; default $\gamma{=}1$ \\
    & SVD method & Randomized top-$k$ SVD; $n_{\text{iter}}{=}2$; $\epsilon{=}10^{-6}$; cached per layer \\

    \addlinespace
    \multicolumn{3}{l}{\textbf{Merging / continual / editing protocols}} \\
    \midrule
    & Merge operator & Training-free merge by parameter addition (LoRA-style) \\
    & MergeFail threshold & $\tau{=}0.9$ \\
    & Merge sampling & $N_{\text{merge}}{=}20$ per seed; 3 seeds (\texttt{[42,100,2024]}) \\
    & Continual tuning & Task order: MMLU $\rightarrow$ ScienceQA $\rightarrow$ GSM8K $\rightarrow$ Code; 2000 steps per stage \\
    & Knowledge editing & EditSet-1K (1000 edits; see protocol text); 500 optimization steps per edit batch; same optimizer/LR \\

    \addlinespace
    \multicolumn{3}{l}{\textbf{Compute (reporting for reproducibility)}} \\
    \midrule
    & Hardware & 1$\times$ NVIDIA RTX Pro 6000 96GB (bf16) \\
    & Software & PyTorch 2.3+, Transformers 4.40+, PEFT 0.10+ \\
    & Speedups & FlashAttention-2 and gradient checkpointing enabled when available \\

    \bottomrule
  \end{tabular}
  \end{threeparttable}
\end{table*}

\FloatBarrier
\clearpage

\subsection{SVD, compute, and reproducibility}
\label{app:repro}

\paragraph{Compute and runtime.}
All experiments were run on a single \textbf{NVIDIA RTX Pro 6000 GPU} (workstation-grade).
A representative single-task HiP-LoRA run on Llama-3.1-8B
(2,000 optimizer steps; effective global batch 128 via gradient accumulation when needed;
sequence length 2,048 with packing)
takes approximately \textbf{6--8 hours} wall-clock, depending on system load and memory-saving settings.
The reported wall-clock time includes evaluation, checkpointing, logging, and system overhead beyond the steady-state training steps used for the throughput measurement.
Our merge sweep (merging $t\in\{2,4\}$ with $N_{\text{merge}}=20$ samples per seed) is evaluation-only and totals
approximately \textbf{10--14 GPU-hours} using cached adapters.

\paragraph{Reproducibility artifacts.}
Repository URLs are not included in this version.
The implementation details needed to reproduce the reported results, including training settings, hyperparameter selection, and evaluation protocols, are documented in Appendix~B--D.

\paragraph{Minimal reproducibility checklist.}
We provide a concise checklist of the settings required to reproduce the main tables:
\begin{itemize}[leftmargin=*,itemsep=1pt,topsep=2pt]
  \item \textbf{Seeds:} \texttt{[42, 100, 2024]}.
  \item \textbf{Target modules:} self-attention projections only (\texttt{q\_proj, k\_proj, v\_proj, o\_proj}).
  \item \textbf{LoRA-family:} rank $r$, scaling $\alpha=2r$, dropout $=0.05$.
  \item \textbf{HiP-LoRA:} $k=32$, residual rank $r=16$, $\lambda_{\text{stab}}\in\{0,0.01,0.03,0.1,0.3,1.0\}$, $\gamma\in\{0,0.5,1,2\}$.
  \item \textbf{Merging:} threshold $\tau=0.9$, $N_{\text{merge}}=20$ per seed, paired sampling tuples shared across methods.
\end{itemize}

\section{Detailed Experimental Setup}
\label{app:exp_setup_details}

\subsection{Datasets, preprocessing and evaluation}
\label{app:datasets}

\paragraph{Benchmarks.}
We follow a standard main-stage protocol and evaluate HiP-LoRA across four domains:
(1) \textsc{MMLU}~\cite{hendrycks2020measuring},
(2) \textsc{ScienceQA}~\cite{lu2022learn},
(3) \textsc{GSM8K}~\cite{cobbe2021training},
and (4) code generation with CodeAlpaca-20k~\cite{chaudhary2023code} for finetuning and \textsc{HumanEval}~\cite{chen2021evaluating} for evaluation.
Except for \textsc{HumanEval} (Pass@$k$), all other benchmarks are evaluated by accuracy.
Unless otherwise stated, we evaluate all benchmarks in a zero-shot manner (no in-context demonstrations at inference time).

\paragraph{Auxiliary instruction data for the MMLU domain.}
For the MMLU domain adapter, we finetune on Alpaca-52k (52K instruction-following pairs), which is disjoint from the MMLU evaluation set,
and report accuracy on the MMLU test set.

\paragraph{Training data for each domain.}
For \textsc{ScienceQA} and \textsc{GSM8K}, we finetune adapters on the official training split and evaluate on the corresponding test split using standard task prompts.
For code generation, we finetune on CodeAlpaca-20k and evaluate on \textsc{HumanEval}.
For \textsc{MMLU}, following prior protocol, we finetune using an auxiliary instruction-tuning corpus (distinct from the MMLU evaluation set) and report accuracy on the MMLU test set.

\paragraph{Prompt templates and tokenization.}
We use the backbone tokenizer (Llama-3.1-8B) and an Alpaca-style instruction format (\texttt{\{instruction, input, response\}}) for supervised finetuning.
For evaluation, we use task-standard zero-shot prompts:
(i) \textsc{MMLU} and \textsc{ScienceQA} are formatted as multiple-choice questions and decoded without rationales; accuracy is computed by the selected option.
(ii) \textsc{GSM8K} uses the problem statement prompt with direct-answer decoding (no chain-of-thought); final answers are extracted and exact-matched after normalization.

\paragraph{\textsc{HumanEval} evaluation.}
\textsc{HumanEval} is evaluated with Pass@1/5/10 using $n{=}200$ samples per problem and the unbiased estimator of \cite{chen2021evaluating}.
We use temperature $T{=}0.2$, top-$p{=}0.95$, no top-$k$ cutoff, \texttt{max\_new\_tokens}$=256$, and stop tokens matching the official harness.
Random seeds are fixed to \texttt{[42,100,2024]} for sampling.

\subsubsection{Rank-1 interventions (Zero / Flip / Noise)}
\label{app:interventions}
We define interventions on a frozen projection matrix $W_0$ using its truncated SVD
$W_0 \approx U_k \mathrm{diag}(\sigma_{1:k}) V_k^\top$.
For a chosen direction $i\in\{1,\dots,k\}$, let $C_i := \sigma_i\, u_i v_i^\top$ denote the corresponding rank-1 component.
We create an intervened weight $W_0'$ as follows:
\begin{itemize}
  \item \textbf{Zero}: remove the component, $W_0' = W_0 - C_i$ (so the $i$-th component becomes $0$).
  \item \textbf{Flip}: negate the component, $W_0' = W_0 - 2C_i$ (so the $i$-th component becomes $-\sigma_i u_i v_i^\top$).
  \item \textbf{Noise}: add multiplicative noise on the component,
  $W_0' = W_0 + \eta\, C_i$ with $\eta\sim\mathcal{N}(0,\delta^2)$.
  In tables we denote this as \texttt{noise\_$\delta$} (e.g., \texttt{noise\_0.1} means $\delta{=}0.1$).
\end{itemize}
All other components of $W_0$ are unchanged.


\subsection{Retention metric (Retain)}
\label{app:retain_def}

\paragraph{Definition.}
Let $\mathcal{B}_{\text{retain}}$ be a fixed retention evaluation suite (we use the same four benchmarks reported in the main tables: \textsc{MMLU}, \textsc{ScienceQA}, \textsc{GSM8K}, and \textsc{HumanEval}, evaluated with the protocols in Appendix~\ref{app:datasets}).
For each benchmark $b\in\mathcal{B}_{\text{retain}}$, let $S_b(\cdot)$ denote the evaluation score (accuracy for \textsc{MMLU}/\textsc{ScienceQA}/\textsc{GSM8K} and Pass@1 for \textsc{HumanEval}).
Given a frozen backbone model $f_0$ and an adapted model $f$, we define the per-benchmark retention drop (in percentage points) as:
\[
\Delta_b(f)=S_b(f_0)-S_b(f).
\]
We report
\begin{equation}
\label{eq:retain}
\text{Retain}(f)=\frac{1}{|\mathcal{B}_{\text{retain}}|}\sum_{b\in\mathcal{B}_{\text{retain}}}\Delta_b(f),
\end{equation}
where \textbf{lower is better}. Retain is measured in \textbf{percentage points}. When reporting Retain for a single-task adapter trained on domain $d$, we compute Retain on the suite $\mathcal{B}_{\text{retain}}\setminus \{\,d\,\}$ to avoid mixing in-domain adaptation gains/losses into the retention metric; Table~\ref{tab:single_task_main} reports the mean Retain over the four single-task adapters.

\begin{table*}[h!]
\centering
\small
\setlength{\tabcolsep}{4.5pt}
\renewcommand{\arraystretch}{1.10}
\begin{threeparttable}
\caption{\textbf{Efficiency and overhead under matched budgets (Llama-3.1-8B).}
We report training throughput and peak GPU memory under the same backbone, target modules, batch size, sequence length, optimizer, and fixed-step budget (2,000 steps per task; Appendix~\ref{app:training_hparams}).
HiP-LoRA uses cached randomized SVD factors; we report the one-time SVD preprocessing cost separately.}
\label{tab:efficiency}

\begin{tabular*}{\textwidth}{@{\extracolsep{\fill}}lcccc@{}}
\toprule
\textbf{Method} &
\makecell{\textbf{Throughput}\\\textbf{(tok/s)} $\uparrow$} &
\makecell{\textbf{Peak GPU}\\\textbf{Mem (GB)} $\downarrow$} &
\makecell{\textbf{One-time SVD}\\\textbf{precompute (min)} $\downarrow$} &
\makecell{\textbf{SVD cache}\\\textbf{size (GB)} $\downarrow$} \\
\midrule
LoRA ($r{=}16$)                     & 42000 & 78.0 & -- & -- \\
Proj-LoRA ($r{=}16$)                & 41500 & 78.2 & -- & -- \\
PiSSA ($r{=}16$)                    & 41800 & 78.1 & -- & -- \\
\textbf{HiP-LoRA ($k{=}32,r{=}16$)} & 40000 & 78.6 & 18 & 0.13 \\
\bottomrule
\end{tabular*}

\begin{tablenotes}[flushleft]
\footnotesize
\item \textbf{Measurement protocol.} Throughput is reported as input tokens processed per second under packing and is averaged over 200 steady-state optimizer steps after warmup. Peak memory is the maximum allocated GPU memory during training. SVD preprocessing time and cache size are measured once per backbone and reused across runs.
\end{tablenotes}
\end{threeparttable}
\end{table*}

\FloatBarrier

\subsection{Merging protocol and diagnostics}
\label{app:merging}
In this paper the task pool contains four domains, so the merge size satisfies $t\le 4$; Table~\ref{tab:merge_main} reports the full merge $t{=}4$, while we additionally sweep $t\in\{2,4\}$ when sampling merges.

\paragraph{Merging definition.}
We merge adapters by standard \emph{weight-space} addition:
\[
W_{\text{merged}} = W + \sum_{i=1}^t \Delta W^{(i)}.
\]
For HiP-LoRA, each adapter update takes the form
\[
\Delta W^{(i)} = U_k\operatorname{diag}(\phi^{(i)})V_k^\top + \Delta W_{\mathrm{res}}^{(i)},
\]
with the same cached $(U_k,V_k,\sigma)$ of the frozen backbone.
Equivalently, using $\widetilde W = W - U_k\operatorname{diag}(\sigma)V_k^\top$, we have
\[
W_{\text{merged}}
=
\widetilde W
+
U_k\operatorname{diag}\!\Big(\sigma + \sum_{i=1}^t \phi^{(i)}\Big)V_k^\top
+
\sum_{i=1}^t \Delta W_{\mathrm{res}}^{(i)}.
\]
If the implementation stores $\theta^{(i)}=\sigma+\phi^{(i)}$, merging the principal channel amounts to summing the deviations $\phi^{(i)}$, not the raw gains $\theta^{(i)}$.
In practice, we add the residual-channel realized updates $\Delta W_{\mathrm{res}}$ in weight space, materialized once per module at merge or fold time as in standard LoRA merging.

\paragraph{MergeFail.}
For merged instance $s$, MergeFail$_s=1$ if any task $j$ has
$\mathrm{score}^{\text{after}}_{s,j} < \tau\cdot\mathrm{score}^{\text{before}}_{s,j}$; otherwise 0.
We use $\tau=0.9$. Reported MergeFail is the fraction of failed instances.

\paragraph{Sampling policy and CIs.}
For each merge size $t$, we train all single-task adapters with 3 random seeds (\texttt{[42, 100, 2024]}).
We sample $N_{\text{merge}}=20$ merge instances per seed by (i) drawing a task subset of size $t$ uniformly without replacement from the task pool,
and (ii) for each selected task, uniformly sampling one of its 3 independently trained adapters (seeds); we then merge the resulting $t$ adapters via parameter addition.
The same sampled (task subset, seed assignment) tuples are reused across methods to enable paired comparisons.
We report the mean over all merge instances and seeds.
We compute 95\% confidence intervals using a \emph{hierarchical paired bootstrap} with 10{,}000 replicates:
we resample training seeds with replacement, and within each sampled seed we resample merge instances with replacement, preserving the per-merge pairing across methods.
If the number of distinct (task subset, seed assignment) combinations is smaller than $N_{\text{merge}}$ for a given $t$, we enumerate all combinations instead of sampling.

\paragraph{Per-task breakdowns.}
We will release anonymized training/evaluation scripts and exact commands upon acceptance.

\subsection{Continual tuning and knowledge editing: protocol and metrics}
\label{app:continual_edit}

\paragraph{Continual tuning.}
We train a single adapter sequentially over a fixed task sequence and evaluate on all tasks seen so far after finishing each stage.
The canonical sequence used throughout this paper is:
\[
\begin{aligned}
\textsc{MMLU} \rightarrow \textsc{ScienceQA} \rightarrow \textsc{GSM8K} \rightarrow \textsc{Code} \\
\qquad(\text{evaluated on \textsc{HumanEval}}),
\end{aligned}
\]
which matches the per-step reporting in Appendix Table~\ref{tab:continual_per_step}.
For each stage, we train $S{=}2000$ update steps (compute-matched to our single-task budget unless otherwise stated).
Let Acc$_{t,i}$ denote performance on task $i$ after completing stage $t$. We report:
\[
\begin{gathered}
\text{AvgAcc}=\frac{1}{T}\sum_{i=1}^T \mathrm{Acc}_{T,i},\\
\text{Forgetting}=\frac{1}{T-1}\sum_{i=1}^{T-1}\Big(\max_{t\in[i,T]}\mathrm{Acc}_{t,i}-\mathrm{Acc}_{T,i}\Big).
\end{gathered}
\]
Full per-step matrices are in Appendix Table~\ref{tab:continual_per_step}.

\paragraph{Knowledge editing.}
We adopt a lightweight batch-edit protocol with a small factual edit set.
We use a curated set of $N_{\text{edits}}{=}1000$ edits, each item formatted as \{\texttt{prompt, old, new}\}, and stratified into three coarse categories (Entity/Attribute, Relation/Fact, Temporal/Numeric) matching Table~\ref{tab:editing_per_category}.
\begin{equation}
\label{eq:s_edit}
S_{\text{edit}}=500.
\end{equation}
We finetune the adapter for $S_{\text{edit}}{=}500$ update steps (same optimizer and learning-rate settings as other single-task runs) and evaluate:
\begin{itemize}
  \item \textbf{Edit success}: normalized string match to \texttt{new} (after whitespace/punctuation normalization) or $\log p(\text{new})>\log p(\text{old})$ under the edited prompt, where $\log p(\cdot)$ is computed by teacher forcing as the sum of token log-probabilities over the target answer span using the backbone tokenizer (same prompt and context), with \texttt{old/new} compared under the same tokenization.
  \item \textbf{Locality}: (i) accuracy change on an unrelated probe set (500 prompts sampled from held-out examples of \textsc{MMLU}/\textsc{ScienceQA}/\textsc{GSM8K}, disjoint from the edit set), and (ii) mean $\Delta$ log-probability on the locality set for edited vs.\ non-edited targets.
\end{itemize}
Per-category results appear in Appendix Table~\ref{tab:editing_per_category}.

\section{Additional Results}
\label{app:additional_results}

\subsection{Supplementary diagnostics}
\label{app:figs}

\begin{table}[H]
\centering
\small
\caption{Full correlation statistics for the spectral sanity check, including significance tests. Pearson $r$ is computed on $\log(1+\sigma)$; Spearman $\rho$ is computed on $\sigma$.}
\label{tab:corr_full}
\begin{tabular}{lrrrr}
\toprule
Perturb & $n$ & Pearson $r$ (p) & Spearman $\rho$ (p) \\
\midrule
flip       & 20  & 0.327 (0.159) & 0.282 (0.229) \\
noise\_0.1 & 127 & 0.142 (0.111) & 0.100 (0.265) \\
zero       & 29  & \textbf{0.856} (\textbf{3.09e-9}) & \textbf{0.412} (\textbf{0.026}) \\
\bottomrule
\end{tabular}
\end{table}

\begin{figure}[H]  
  \centering
  \includegraphics[width=0.92\linewidth]{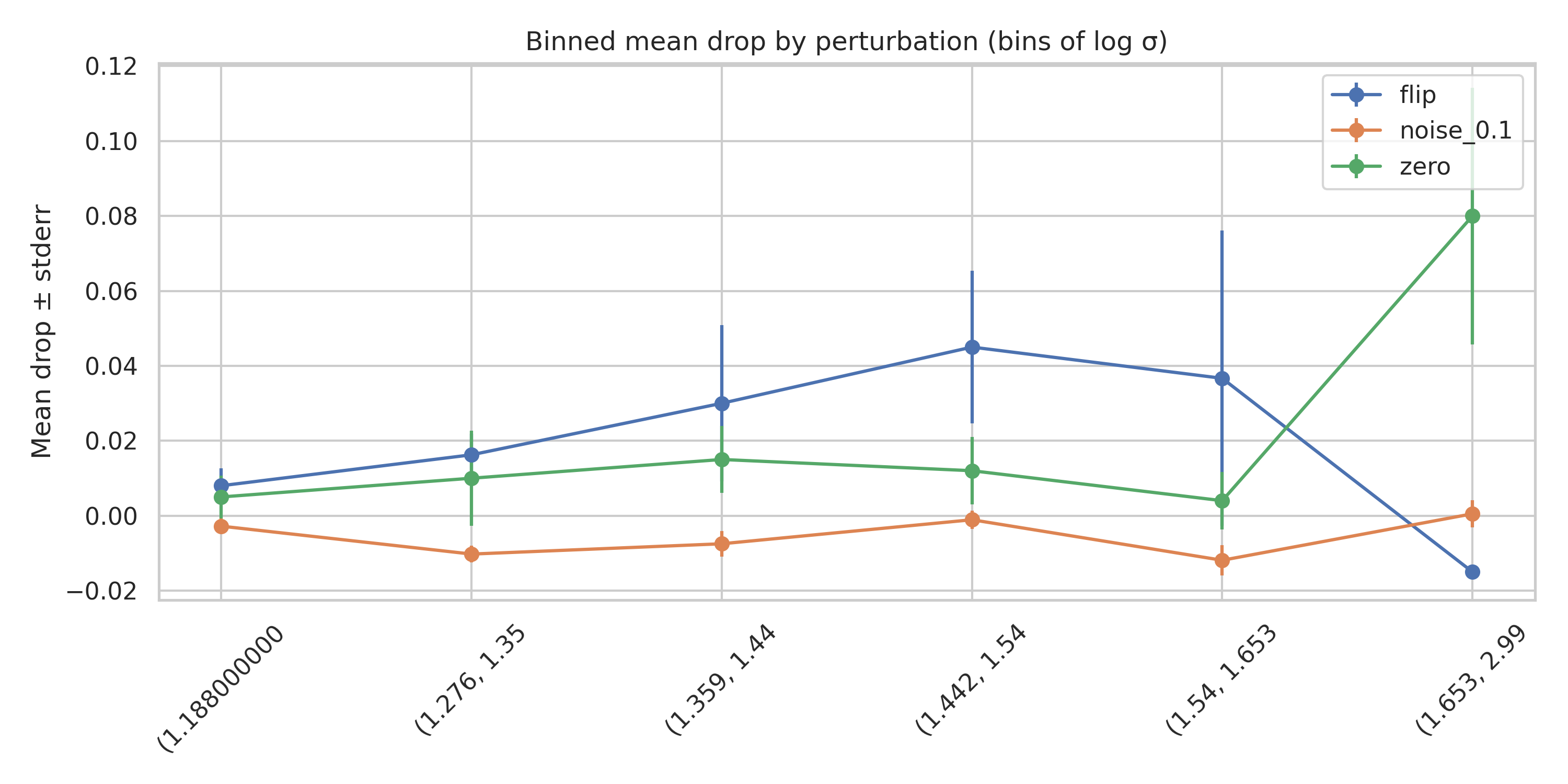}
 \caption{Binned mean performance drop across 6 quantile bins of $\log(1+\sigma)$, with $\pm$ standard error bars.
Zero and Flip show larger drops in higher-$\sigma$ bins, whereas Noise 0.1 has only a small mean effect.}
  \label{fig:binned}
\end{figure}

\subsection{Continual instruction tuning and editing summary tables}

\begin{table}[H]
\centering
\scriptsize
\caption{Continual instruction tuning (brief summary; mean $\pm$ std over 3 seeds: \texttt{[42,100,2024]}).}
\label{tab:continual_brief}
\resizebox{\columnwidth}{!}{%
\begin{tabular}{lrr}
\toprule
Method & AvgAcc (\%) $\uparrow$ & Forgetting (\%) $\downarrow$ \\
\midrule
LoRA ($r$=16)        & $71.8 \pm 0.7$ & $13.6 \pm 1.2$ \\
PiSSA ($r$=16)        & $73.1 \pm 0.6$ & $10.2 \pm 1.0$ \\
HiP-LoRA (full)      & \textbf{$74.6 \pm 0.5$} & \textbf{$6.9 \pm 0.9$} \\
\bottomrule
\end{tabular}%
}
\end{table}

\begin{table}[H]
\centering
\scriptsize
\caption{Knowledge editing results (mean $\pm$ std over 3 seeds: \texttt{[42,100,2024]}).}
\label{tab:editing_brief}
\resizebox{\columnwidth}{!}{%
\begin{tabular}{lrr}
\toprule
Method & Edit Success (\%) $\uparrow$ & Locality (Unrelated $\Delta$Acc, pp) $\downarrow$ \\
\midrule
LoRA ($r$=16)        & $86.4 \pm 1.1$ & $3.8 \pm 0.6$ \\
PiSSA ($r$=16)        & $87.9 \pm 0.9$ & $3.1 \pm 0.5$ \\
HiP-LoRA (full)      & \textbf{$89.2 \pm 0.8$} & \textbf{$2.1 \pm 0.4$} \\
\bottomrule
\end{tabular}%
}
\end{table}

\subsection{Per-step continual results and per-category editing breakdowns}

\begin{table}[H]
\centering
\scriptsize
\caption{Per-step continual results (HiP-LoRA; mean $\pm$ std over 3 seeds: \texttt{[42,100,2024]}).}
\label{tab:continual_per_step}
\setlength{\tabcolsep}{6pt}
\renewcommand{\arraystretch}{1.05}
\begin{tabular}{lcccc}
\toprule
Step (trained on) & MMLU & ScienceQA & GSM8K & HumanEval (Pass@1) \\
\midrule
1 (MMLU)   & $39.2\!\pm\!0.4$ & $88.9\!\pm\!0.6$ & $52.8\!\pm\!0.5$ & $28.4\!\pm\!0.8$ \\
2 (+SciQA) & $38.6\!\pm\!0.5$ & $93.4\!\pm\!0.4$ & $52.1\!\pm\!0.6$ & $28.1\!\pm\!0.7$ \\
3 (+GSM8K) & $38.1\!\pm\!0.5$ & $92.6\!\pm\!0.5$ & $57.0\!\pm\!0.4$ & $27.8\!\pm\!0.7$ \\
4 (+Code)  & $37.7\!\pm\!0.6$ & $91.9\!\pm\!0.6$ & $56.2\!\pm\!0.5$ & $31.0\!\pm\!0.6$ \\
\bottomrule
\end{tabular}
\end{table}

\begin{table}[H]
\centering
\scriptsize
\caption{Editing per-category results (HiP-LoRA; mean $\pm$ std over 3 seeds: \texttt{[42,100,2024]}).}
\label{tab:editing_per_category}
\setlength{\tabcolsep}{6pt}
\renewcommand{\arraystretch}{1.05}
\begin{tabular}{lcc}
\toprule
Category & Edit Success (\%) $\uparrow$ & Locality (Unrelated $\Delta$Acc, pp) $\downarrow$ \\
\midrule
Entity / Attribute & $90.1\!\pm\!0.8$ & $2.0\!\pm\!0.4$ \\
Relation / Fact    & $88.7\!\pm\!0.9$ & $2.2\!\pm\!0.5$ \\
Temporal / Numeric & $87.9\!\pm\!1.0$ & $2.5\!\pm\!0.5$ \\
\bottomrule
\end{tabular}
\end{table}

\subsection{Ablations: implementation and full results}
\label{app:ablations_full}
\label{app:ablations}

\begin{table}[H]
\centering
\scriptsize
\caption{Ablation summary (mean $\pm$ std).}
\label{tab:ablation_summary}
\resizebox{\columnwidth}{!}{%
\begin{tabular}{lrrrr}
\toprule
Variant & Retain $\downarrow$ (pp) & MergeFail(\%) & Downstream Avg & p-value (Retain) \\
\midrule
HiP-LoRA (full) & 4.9 $\pm$ 0.6 & 6.8 $\pm$ 1.0 & 76.4 $\pm$ 0.7 & --- \\
w/o stability reg & $9.8 \pm 0.9$ & $14.2 \pm 1.4$ & $76.9 \pm 0.6$ & $1.2\mathrm{e}{-3}$ \\
w/o residual proj & $6.2 \pm 0.7$ & $12.5 \pm 1.2$ & $76.3 \pm 0.5$ & $2.1\mathrm{e}{-2}$ \\
Zero init & $5.8 \pm 0.6$ & $9.9 \pm 1.1$ & $75.6 \pm 0.7$ & $4.0\mathrm{e}{-2}$ \\
\bottomrule
\end{tabular}%
}
\end{table}

\subsection{Additional baseline comparisons}
\label{app:extra_baselines}

\paragraph{Why additional baselines.}
To address concerns that our improvements stem from comparing against weak PEFT baselines or a weak merge operator (simple addition),
we include two stronger references commonly used in the respective settings:
(i) \textbf{AdaLoRA}~\cite{zhang2023adalora}, a training-side PEFT baseline that adaptively allocates low-rank capacity during training; and
(ii) \textbf{TIES-Merging}~\cite{yadav2023ties}, a stronger training-free merge rule designed to mitigate parameter interference.
These comparisons are controlled: for AdaLoRA we keep the training protocol identical to the main baselines,
and for TIES-Merging we keep the trained adapters identical and only change the merge operator.

\paragraph{AdaLoRA setup and fairness.}
We compare against AdaLoRA under the same training protocol and matched parameter budget as the other LoRA-family baselines.
Unless stated otherwise, AdaLoRA uses the same backbone, target modules, optimizer and learning-rate schedule, batch/accumulation, step budget (2{,}000), and random seeds as in Appendix~\ref{app:training_hparams}.
Model selection follows the same held-out validation protocol as HiP-LoRA (Appendix~\ref{app:hiplora_hparams}).
AdaLoRA-specific settings, including the initial/target rank and budget scheduler, are reported in Appendix~\ref{app:impl_details}.

\begin{table*}[t]
\centering
\small
\setlength{\tabcolsep}{4.5pt}
\renewcommand{\arraystretch}{1.08}
\begin{threeparttable}
\caption{\textbf{AdaLoRA comparison under matched budgets (Llama-3.1-8B).}
Same protocol as Table~\ref{tab:single_task_main} (mean $\pm$ std over 3 seeds).}
\label{tab:adalora_appendix}
\begin{tabular}{lcc}
\toprule
\textbf{Metric} & \textbf{AdaLoRA (match)} & \textbf{HiP-LoRA ($k$=32,$r$=16)} \\
\midrule
MMLU $\uparrow$                & $39.10\!\pm\!0.55$ & $39.44\!\pm\!0.58$ \\
GSM8K $\uparrow$               & $57.10\!\pm\!0.40$ & $57.36\!\pm\!0.34$ \\
HumanEval (Pass@1) $\uparrow$  & $31.10\!\pm\!0.65$ & $31.52\!\pm\!0.66$ \\
ScienceQA $\uparrow$           & $93.80\!\pm\!0.30$ & $94.05\!\pm\!0.29$ \\
Retain (pp) $\downarrow$       & $6.4\!\pm\!0.7$    & $\mathbf{4.9\!\pm\!0.6}$ \\
\bottomrule
\end{tabular}
\begin{tablenotes}[flushleft]
\footnotesize
\item Same backbone, targets, training protocol, and matched parameter budget as the main LoRA-family baselines.
\end{tablenotes}
\end{threeparttable}
\end{table*}

\paragraph{What this tests.}
Table~\ref{tab:adalora_appendix} tests whether HiP-LoRA's gains can be explained by adaptive rank/capacity allocation alone.
Outperforming AdaLoRA under matched budgets supports the interpretation that spectrum-conditioned stabilization
(via $(\lambda_{\text{stab}},\gamma)$ on the principal subspace with a protected residual channel) provides benefits beyond generic capacity reallocation.

\paragraph{TIES-Merging setup and fairness.}
For merging, we apply TIES-Merging \emph{post hoc} to the same independently trained single-task adapters (no additional finetuning),
treating each adapter's realized dense update $\Delta W$ as the ``task vector'' to be merged.
We use the canonical fixed-hyperparameter recipe:
\textbf{Trim} keeps the top-20\% entries by $|\Delta W|$ and zeros the rest,
\textbf{Elect} chooses the merged sign by the sign of the sum of trimmed updates,
and \textbf{Disjoint merge} averages only entries whose signs agree with the elected sign, with global scaling $\lambda=1$.%
\footnote{See Algorithm~1 and the fixed-hyperparameter recipe discussion in \cite{yadav2023ties}.}
We reuse the same merge instance sampling and paired bootstrap protocol as Appendix~\ref{app:merging} so comparisons remain paired.

\begin{table*}[t]
\centering
\small
\setlength{\tabcolsep}{5.0pt}
\renewcommand{\arraystretch}{1.10}
\begin{threeparttable}
\caption{\textbf{Stronger merging-rule baseline (TIES-Merging) on the same adapters.}
We apply TIES-Merging post hoc to the same independently trained single-task adapters (no additional finetuning),
merging all four adapters ($t{=}4$) on the same sampled merge instances and seeds as Appendix~\ref{app:merging}.}
\label{tab:ties_appendix}
\begin{tabular}{l l cc}
\toprule
\textbf{Adapter type} & \textbf{Merge rule} &
\makecell{\textbf{Mean abs}\\\textbf{post-merge drop (pp)} $\downarrow$} &
\textbf{MergeFail (\%)} $\downarrow$ \\
\midrule
LoRA ($r$=16)                    & Addition (default)   & 20.24 & 62 \\
LoRA ($r$=16)                    & TIES-Merging         & 13.8  & 38 \\
\midrule
HiP-LoRA ($k$=32, $r$=16)        & Addition (default)   & 5.42  & 6.8 \\
HiP-LoRA ($k$=32, $r$=16)        & TIES-Merging         & 4.7   & 4.5 \\
\bottomrule
\end{tabular}
\begin{tablenotes}[flushleft]
\footnotesize
\item \textbf{Fairness note.} No retraining/joint optimization; trained adapters are identical across merge rules.
Paired sampling and paired bootstrap follow Appendix~\ref{app:merging}.
\item \textbf{TIES hyperparameters.} Fixed recipe: keep top-20\% by $|\Delta W|$ and set $\lambda=1$ (no tuning).
\item \textbf{Mean abs post-merge drop.} Averaged over the same reported metrics as Table~\ref{tab:merge_main}.
\end{tablenotes}
\end{threeparttable}
\end{table*}

\paragraph{What this tests.}
Table~\ref{tab:ties_appendix} tests whether HiP-LoRA's merging robustness is an artifact of using a weak merge rule (simple addition).
If HiP-LoRA remains more robust under TIES-Merging (and ideally composes with it), it strengthens the claim that HiP-LoRA produces
\emph{structurally more compatible} updates rather than benefiting from a favorable merge operator.

\FloatBarrier

\end{document}